\documentclass[letterpaper]{article} 
\usepackage{aaai23}  
\usepackage{times}  
\usepackage{helvet}  
\usepackage{courier}  
\usepackage[hyphens]{url}  
\usepackage{graphicx} 
\usepackage{natbib}  
\usepackage{caption} 
\usepackage{algorithm}
\usepackage{algorithmic}
\usepackage[none]{hyphenat}
\usepackage{newfloat}
\usepackage{listings}
\DeclareCaptionStyle{ruled}{labelfont=normalfont,labelsep=colon,strut=off} 
\floatstyle{ruled}
\newfloat{listing}{tb}{lst}{}
\floatname{listing}{Listing}
\usepackage{xcolor}

\title{Integrating Curricula with Replays: Its Effects on Continual Learning}

\author {
    Tee Ren Jie,\textsuperscript{\rm 1,\rm 2}
    Mengmi Zhang \textsuperscript{\rm 2, \rm 3}
}
\affiliations {
    \textsuperscript{\rm 1} Hwa Chong Institution, Singapore\\
    \textsuperscript{\rm 2} CFAR and I2R, Agency for Science, Technology, and Research, Singapore\\
    \textsuperscript{\rm 3} Nanyang Technological University, Singapore\\
    Address correspondence to mengmi@i2r.a-star.edu.sg
}

\begin{document}

\maketitle


\begin{abstract}
Humans engage in learning and reviewing processes with curricula when acquiring new skills or knowledge. 
This human learning behavior has inspired the integration of curricula with replay methods in continual learning agents. The goal is to emulate the human learning process, thereby improving knowledge retention and facilitating learning transfer. 
Existing replay methods in continual learning agents involve the random selection and ordering of data from previous tasks, which has shown to be effective. However, limited research has explored the integration of different curricula with replay methods to enhance continual learning.
Our study takes initial steps in examining the impact of integrating curricula  with replay methods on continual learning in three specific aspects: the interleaved frequency of replayed exemplars with training data, the sequence in which exemplars are replayed, and the strategy for selecting exemplars into the replay buffer. These aspects of curricula design align with cognitive psychology principles and leverage the benefits of interleaved practice during replays, easy-to-hard rehearsal, and exemplar selection strategy involving exemplars from a uniform distribution of difficulties.
Based on our results, these three curricula 
effectively mitigated catastrophic forgetting and enhanced positive knowledge transfer, demonstrating the potential of curricula in advancing continual learning methodologies. Our code and data are available: \url{https://github.com/ZhangLab-DeepNeuroCogLab/Integrating-Curricula-with-Replays}

\end{abstract}

\section{Introduction}



Continual learning enables consecutive task acquisition without forgetting previously trained tasks \cite{Kirkpatrick2017}. This adaptability is vital for autonomous systems in dynamic environments, such as updating a grocery classification model with new products without retraining it on previous products. However, a significant challenge in continual learning is catastrophic forgetting, where knowledge from recent tasks interferes with earlier ones \cite{Parisi2019}, leading to performance degradation on earlier tasks after training on a task sequence.

To resolve this problem, 
there are three primary types of continual learning methods commonly employed in the field: 
regularization-based methods introduce regularization terms to mitigate catastrophic forgetting by preserving important parameters during training \cite{chaudhry2018riemannian, zenke2017continual, aljundi2018memory, benzing2022unifying}; rehearsal-based methods store and replay a subset of previous data during training to maintain knowledge from previous tasks \cite{rolnick2019experience, chaudhry2019tiny, riemer2018learning, vitter1985random, rebuffi2017icarl, castro2018end} and parameter isolation methods isolate specific parameters for each task to prevent interference between tasks \cite{yoon2017lifelong, hung2019compacting, ostapenko2019learning}. 
Rehearsal-based methods have proven highly effective in continual learning. However, existing approaches typically involve randomly selecting and rehearsing data from previous tasks. Limited research explores the incorporation of meaningful curricula into replay methods.

In parallel, in the curriculum learning literature, various approaches have focused on weakly supervised \cite{guo2018curriculumnet}, unsupervised \cite{azad2023clutr}, and reinforcement learning tasks \cite{10.5555/3455716.3455897}. These studies demonstrate that curricula improve generalization abilities, task performances,
and convergence speed \cite{wu2021when, Soviany2022} during training. However, they primarily address intra-class difficulty and example scheduling within a single task, neglecting the impact of class presentation sequences across multiple tasks. Recent research has explored curricula in continual learning scenarios without data replays \cite{learingtolearn}. In complement to this work, our study investigates the role of curricula specifically during replay in continual learning, while keeping the curricula consistent for the feed-forward training process.




Exploring optimal curricula offers countless possibilities, and in our study, we take initial steps to investigate a limited set of potential curricula. We draw inspiration from two sources to guide the design of these curricula. Firstly, neuroscience research has revealed that neural activity patterns associated with past experiences are replayed in specific orders during rest or sleep, which is believed to contribute to memory consolidation and spatial navigation \cite{Drieu2019}. Secondly, pedagogy studies indicate that repetitive practice and revisiting previous knowledge with increasing difficulty enhance long-term memory integration in students \cite{Zhan2018}. 

Specifically, we propose three types of curricula for replays and examine their impact on catastrophic forgetting and positive knowledge transfer: (1) the interleaved frequency of replayed exemplars with training data, (2) the replay sequence of exemplars, and (3) the strategy for selecting exemplars into the replay buffer. The experimental findings align with cognitive psychology principles, highlighting the advantages of frequently interleaving between training data and replayed exemplars, incorporating easy-to-hard rehearsals, and selecting exemplars from a uniform distribution of difficulties for replay. These observations present a promising avenue for advancing continual learning methods. It also provides insights into the underlying mechanisms of replay strategies in mitigating forgetting and facilitating knowledge transfer across tasks.

\section{Related Works}

\subsection{Replay Methods in Continual Learning}

Extensive research has focused on utilizing replay methods to address the issue of catastrophic forgetting. Conventional replay methods, such as iCaRL \cite{rebuffi2017icarl} and ER \cite{rolnick2019experience}, involve explicit training on previously saved data, while several variants, like DGR \cite{shin2017continual} and Pseudo-Recursal \cite{DBLP:journals/corr/abs-1802-03875}, replay on artificially synthesized samples by generative models, resembling data from previous tasks.
Although these replay methods have made significant contributions in reducing catastrophic forgetting, they paid little attention to the incorporation of meaningful curricula into replay methods. Most methods randomly interleave the replay samples with the training data, without exploring the optimal mixing strategies \cite{rolnick2019experience, Ven2020, DBLP:journals/corr/abs-1903-02647}. In our work, we systematically studied the effect of interleaving curricula, which involves mixing training data and replay samples within a pre-defined interleave interval. 



\subsection{Curriculum Learning}


Curriculum learning methods can be broadly categorized into two groups. The first group involves manual curriculum design by humans before training \cite{9093408, 7410525}, but these methods typically rely on human expertise and struggle to generalize to new domains. The second group consists of models that can autonomously design curricula without human intervention \cite{pmlr-v70-graves17a, DBLP:journals/corr/abs-1805-03643}. However, the application of these methods to enhance model performance has received limited attention in the continual learning setting. 
Here, we highlight two factors to consider when applying curricula on the replay methods in continual learning. Firstly, while curriculum learning has demonstrated efficacy in enhancing generalization and training speed within a single task, the objective of curriculum learning in the context of continual learning is to retain knowledge from previous tasks while acquiring new knowledge from the current task. Secondly, unlike within-task curriculum learning, models in continual learning only have access to data from the current task, making it challenging to create a comprehensive between-task curriculum that encompasses the entire dataset. 
Here, we took initial steps in this direction by exploring automated methods to determine the sequence of replay samples and introducing the sample selection strategy which finds the best replay samples for building a curriculum.

\section{Experiments}

We investigated the effect of three types of replay curricula in the class incremental learning (CIL) setting. We first introduce CIL, and then elaborate on the three replay curricula individually.

\noindent \textbf{Problem Setting.} The objective of CIL is to teach a unified classification model $\Theta$ to recognize sets of object classes incrementally over time. Specifically, an image dataset $D$,  consisting of $N$ object classes, is split into subsets $\{D_1,...,D_t,...,D_T\}$ 
of 
images 
and presented over a sequence of $T$ tasks. In each task $t$, the model only has access to training data in $D_t$, consisting of 
samples from distinct classes $C_t$, and $(x_{i,t},y_{i,t})$ is the $i$-th (image, label) pair in $D_{t}$. The model $\Theta$ can run multiple passes over $D_t$ in task $t$. The model stops training on $D_t$ 
after its performance on the validation set saturates, considering the five most recent epochs.

We implemented the naive replay method where some raw images and their corresponding labels are selected from previous tasks and are stored in the replay buffer $R_t$. These data in $R_t$ are inter-leaved with $D_t$ for rehearsals. There are three types of replay curricula involved in this study: (1) the interleave frequency; (2) the rehearsal sequence of $R_t$ in CIL; and (3) the image selection for $R_t$. 
$R_t$ is kept at a constant size of 1200 over all the tasks. See \textbf{Appendix} for more training details. 
As an upper bound, we also include the \textbf{offline method} where the model $\Theta$ is trained on the entire dataset $\Theta$ from $\{D_1,...,D_T\}$ over multiple epochs without any continual learning.

\noindent \textbf{Datasets.}  We conducted experiments to investigate the use of these three types of curricula in replay methods on the two image datasets ciFAIR-10 and ciFAIR-100 \cite{Barz2019}.
ciFAIR-10 dataset contains 10 object classes. The protocol asks the model $\Theta$ to incrementally learn 2 object classes in each task. There are a total of 5 tasks. ciFAIR-100 dataset contains 100 object classes. The CIL protocol asks the model $\Theta$ to incrementally learn 5 object classes in each task. There are a total of 20 tasks. 
Both datasets have a total of 60,000 images, with 50,000 images used for training and 10,000 images used for testing.
The conclusions drawn from the experiments on both datasets are consistent. Without loss of generality, we focus on all the experiments and result analysis in ciFAIR-100 in the main text.
See \textbf{Appendix} for more implementation details and results on ciFAIR-10.

\noindent \textbf{Evaluation Metrics.} To assess the continual learning performance of the model $\Theta$, we followed \cite{learingtolearn} and introduce 2 standard evaluation metrics. We define \textbf{Forgetfullness (F)} as the percentage decrease in classification accuracy on the test instances from $C_{1}$ 
between the $\Theta_{t}$ (model after being trained on $D_t$) and $\Theta_{1}$. An ideal $\Theta_{t}$ could maintain the same classification accuracy on $C_{1}$ over tasks; i.e. $\forall t, F_t=0$. The higher F is, the more $\Theta$ suffers from catastrophic forgetting. To assess the overall classification performance of $\Theta$ over tasks, we also report the continual average classification accuracy (\textbf{Avg. Accu.}). Avg. Accu. is computed as the average accuracy on all the test instance from $C_i$, where $i\in \{1, 2, ..., t\}$. For simplicity, we report the averaged \textbf{F} and \textbf{Avg. Accu.} over all the tasks.

 \noindent \textbf{Experimental Controls.}
 Within each experiment, only one variable of interest changes while the rest of the experiment conditions are fixed as control variables. As the previous study has shown that the sequence of class presentations affects the continual learning performance \cite{HE202267}, we use the same class presentation sequence in all three experiments. The same MobileNetV3 (small) network architecture is used as the backbone for the model $\Theta$ for all experiments. In every experiment, the total number of training samples and the total number of replay samples exposed to $\Theta$ remain the same across all experiment variables. Each experiment is conducted with 4 runs initialized with 4 random seeds, where the seeds are used to vary the controlled variables. The average performance across all 4 runs is reported. 





\subsection{Interleave Divisions during Rehearsals}

The number of interleaving divisions refers to the number of splits of in $D_t$ and $R_t$. It indicates how often the model $\Theta$ rehearses on $R_t$, while learning on a subset of $D_t$. For example, for interleaving division number 400, $D_t$ is split into 400 groups where each group contains an equal number of $(x_{i,t},y_{i,t})$ (image, label) pairs, and these (image, label) pairs are randomly selected from $D_t$ without replacement. Correspondingly, $R_t$ is also split into 400 groups with the same splitting criteria as $D_t$. At each training epoch, the model $\Theta_t$ at task $t$ is repeatedly trained with one group of $D_t$ followed by one group of $R_t$, until the entire $D_t$ and $R_t$ are exhaustively seen by $\Theta_t$. We titrate the interleave division numbers with the range of 1, 8, 60, 120, and 300. 
The training data is interleaved with replay data and then presented to the model in sequence. Different interleave division numbers result in different data presentation sequences; hence, different curricula.  

\subsection{Rehearsal Sequence of Replay Samples}

We use the interleave divisions 1 and 600 for all the experiments in this subsection and vary the rehearsal sequence of data samples in $R_t$ by taking into account the two factors: the sample difficulty levels and the increasing or decreasing directions of sample difficulty levels.

To measure whether a sample is easy or hard to learn, we introduce two difficulty measures: (1) the confidence score difficulty metrics and (2) the distance vector difficulty metrics. The confidence score difficulty metrics were used to assess whether a teacher network with full knowledge of the entire dataset $D$ predicted high or low confidence of the given sample belonging to its ground truth class label. Specifically, each image within $R_t$ was input to a teacher network. The teacher network is based on a MobileNetV3 (small) architecture, pre-trained on the entire dataset $D$. After this, the confidence score for the ground truth class of each sample was extracted from the teacher network’s output. $R_t$ was then sorted according to its individual sample’s confidence score, where a higher confidence score means that the sample is easier to learn for $\Theta$. 

However, in CIL setting, having a teacher network with full access to the whole dataset is impractical, as the data is incrementally available over tasks. Hence, we employed the distance vector difficulty metrics, used widely in literature \cite{learingtolearn, huang2022curriculum}. Intuitively, if the sample is closer to other samples in the memory buffer, it is easier for $\Theta$ to learn and generalize to other samples as well.
The 2nd last layer from a ResNet-50 model \cite{7780459}, pretrained on the ImageNet dataset, was used to extract the feature vector of each sample in $R_t$. A Euclidean distance matrix was created, where the pairwise Euclidean distance for all the samples based on their feature vectors was computed. We then compute the sum of each row of the matrix and denote this column vector as a distance vector. Each element in this distance vector represents how much a particular sample differs from all other samples in the feature space. A smaller value in the distance vector means that the particular replay sample is easier to learn for $\Theta$. 


We introduce a series of rehearsal sequences in the orders of either easy-to-hard samples or hard-to-easy samples, where the difficulty levels of each sample are determined by either the confidence score difficulty metrics or the distance vector difficulty metrics.
As the previous study has shown that the class orders are also essential for continual learning \cite{HE202267}, here we also explore the effect of the class orders during replays. When we design the rehearsal sequence based on class difficulties in $R_t$, we adapt the two sample-level difficulty measures above to compute class-level difficulty measures by taking the average over all samples of the same class in $R_t$. We then sort all the samples in $R_t$ by their class difficulty metrics, regardless of their individual sample difficulty scores. 
Samples in $R_t$ sorted by their class difficulties
were then presented to the model $\Theta$ in either the easy-to-hard or hard-to-easy 
orders.

\subsection{Selection of Samples for Replay Buffer}

In common practice, selecting samples for $R_{t+1}$ from task $t$ is often conducted in a random manner \cite{rolnick2019experience,bagus2021investigation}. In contrast to the previous works, we vary the sample selection criteria for $R_{t+1}$ as follows: selecting only the easiest samples from task $t$ for $R_{t+1}$, selecting the hardest samples from task $t$ for $R_{t+1}$, and selecting samples that are uniformly distributed across difficulty levels from task $t$ for $R_{t+1}$. The difficulty levels are judged based on the confidence scores and the distance vectors defined in the previous subsection. We use interleave division numbers 1 and 600 for all the experiments in this subsection.


\section{Results}

\subsection{Frequent Replays Enhance Performances}

\begin{table}[t]
\centering
\begin{tabular}{|c|c|c|c|c|c|}
\hline
Inter. Div. & 1    & 8    & 60   & 120  & 300  \\ \hline
F           & 63.9 & 62.6 & 57.4 & \textbf{55.1} & 56.1 \\ \hline
Avg.Accu    & 39.9 & 40.7 & 44.6 & \textbf{46.6} & \textbf{46.6} \\ \hline
\end{tabular}\vspace{-3mm}
\caption{\textbf{Continual learning performance on ciFAIR-100 as a function of interleave divisions (Inter. Div.)}. The higher Avg. Accu., the better; the lower F, the better. The best are in bold. The offline accuracy is 62.0\%. See \textbf{Experiments Section} for evaluation metrics and experiments in interleave divisions.}
\label{tab1:exp1}\vspace{-4mm}
\end{table}

We report \textbf{F} and \textbf{Avg. Accu.} as a function of interleave divisions in Table \ref{tab1:exp1}. 
Notably, we observed that interleave divisions are important factors influencing the continual learning performance of the replay method with the larger interleave divisions leading to better performances, as indicated by the decreasing \textbf{F} and increasing \textbf{Avg. Accu.} over all the tasks. It is possible that the model parameters at large division numbers are updated more frequently for both the current task and all previous tasks, resulting in minimal forgetting. However, we also note that the continual learning performance saturates at interleave division number 120. This implies that increasing interleave divisions beyond optimal values brings no extra benefits in continual learning.

\subsection{Easy-To-Hard Rehearsal Sequences are Beneficial}

\begin{figure}[t]
\centering
\includegraphics[width=8cm]{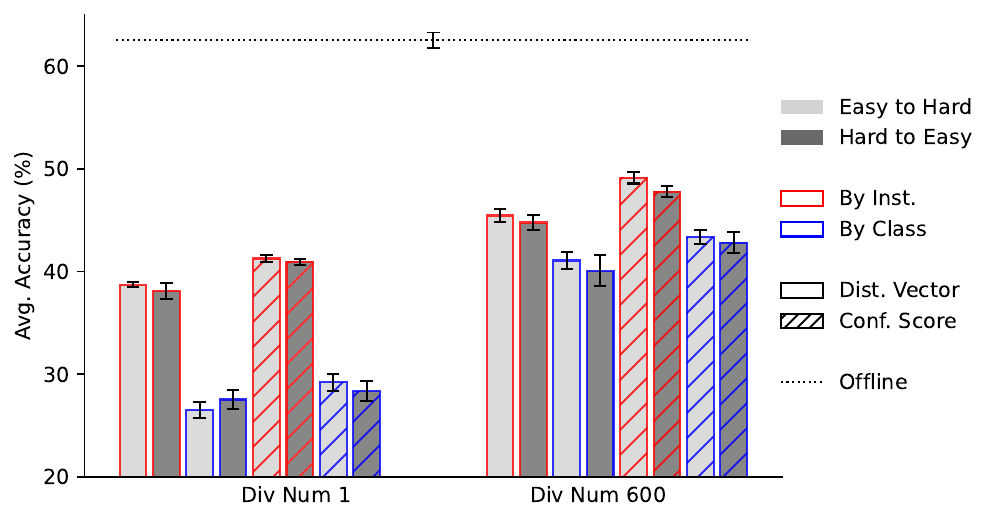}\vspace{-3mm}
\caption{\textbf{Continual learning performance on ciFAIR-100 dataset with different rehearsal sequences.} Average classification accuracy over all tasks for models trained with different rehearsal sequences is reported. The higher the average accuracy, the better. See the legend on the right for conventions of different rehearsal sequence conditions. For example, a light grey bar with a red boundary without the texture of slanted bar patterns (the leftmost bar) indicates that the model is trained with an easy-to-hard rehearsal sequence sorted by the instance-level difficulties based on distance vector metrics. See \textbf{Experiments Section} for details. See \textbf{Appendix} for the results reported in \textbf{F}.
}
\vspace{-4mm}
\label{fig:exp2}    
\end{figure}

We studied the models trained with different rehearsal sequences sorted in easy-to-hard or hard-to-easy curricula based on sample-level or class-level difficulty measures computed from either the confidence scores or distance vectors. We reported the Avg. Accu. results in Figure \ref{fig:exp2} and F scores in \textbf{Appendix} and made four key observations. First, aligning with the observations in Table \ref{tab1:exp1} and the discussion from the previous subsection, large interleave divisions benefit continual learning models with higher average accuracy and less forgetting. Second, rehearsal sequences sorted by instance-level difficulties lead to much better continual learning performances (compare red bars versus blue bars). Third, the confidence score is a better evaluation metric measuring instance-level difficulties, as shown by the bars with and without texture patterns. Finally, the models trained with the easy-to-hard rehearsal sequences outperform the ones with reversed rehearsal sequences (compare light versus dark grey bars). It is possible that easy-to-hard rehearsal sequences make the models converge faster on the previous tasks due to more stable gradient updates; hence, the sequences lead to minimal forgetting and higher classification accuracy. We also compared the continual learning performance for both the offline method and the continual learning method with various curricula and observed that there still exists a large performance gap between these two.     

\subsection{Replays with Only Hard Data Hurt Performances}

\begin{figure}[t]
\centering
\includegraphics[width=8cm]{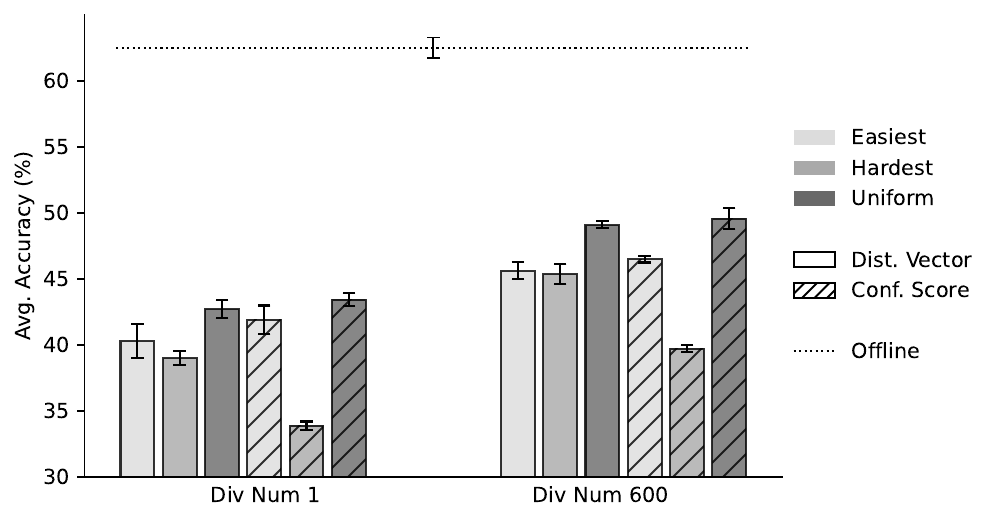}\vspace{-3mm}
\caption{\textbf{Continual learning performance on ciFAIR-100 dataset under different replay sample selection strategies.} Average classification accuracy over all tasks for models trained with different replay samples selected using different strategies. The interpretation of legends and design conventions are similar to Figure \ref{fig:exp2}. 
See \textbf{Experiments Section} for details. See \textbf{Appendix} for the results in \textbf{F}.
}\vspace{-4mm}
\label{fig:exp3}    
\end{figure}

Here, we explored the effect of different sample selection strategies for replay samples in terms of the sample difficulty levels based on distance vectors or confidence scores. From Figure \ref{fig:exp3},
Our observations indicate that exclusively choosing the most challenging replay samples leads to inferior performance compared to selecting the easiest samples or incorporating samples with a balanced distribution of difficulty levels. Selecting samples with a uniform distribution of difficulty levels yields the best continual learning performance. This outcome may be attributed to the fact that difficult replay samples result in less flat loss landscapes, which in turn make the training process more challenging and slower to converge \cite{wu2021when}. A curriculum for training the models to rehearse from the easiest to the hardest samples is the best, as it balances the greater precision in data fitting due to the hardest samples and the fast convergence speed during training due to the easier samples. Similar to the previous subsection, we also noted that the confidence score is a better measure of sample difficulty levels than the distance vectors. 


\section{Conclusion}
Our study 
examines the role of curricula during replays in the class-incremental learning setting in continual learning. We designed and conducted a series of controlled experiments to study the three key questions on replays: how often is the replay, what data should be replayed, and in what sequence to replay.
Across the two common image datasets, our experimental results shed light on the underlying principles of replay methods in continual learning and reveal the good curricula design choices for replay methods. 
These curricula designs not only facilitate positive knowledge transfers (which has been explored in existing curriculum learning literature), but also mitigate catastrophic forgetting (a significant problem we need to solve in continual learning). Specifically, we 
found that (1) replays should happen frequently; (2) only rehearsing on the most difficult exemplars hurts continual learning performances; and (3) rehearsals on samples with increasing difficulty eliminate forgetting more than its reversed difficulty orders. 
There are numerous other possible choices of curricula designs for replay methods, such as a unified difficulty metric considering both confidence scores and distance vectors or the use of a student feedback loop to update the difficulty scores. In the future, we will look into the role of curricula under
stringent continual learning conditions, such as learning with limited training time or noisy data. We will also conduct experiments on other large-scale datasets and apply our replay curriculum to existing replay-based continual learning algorithms. 

\section{Acknowledgements}
This research is supported by the National Research Foundation, Singapore under its AI Singapore Programme (AISG Award No: AISG2-RP-2021-025), its NRFF award NRF-NRFF15-2023-0001, Mengmi Zhang's Startup Grant from Agency for Science, Technology, and Research (A*STAR), and Early Career Investigatorship from Center for Frontier AI Research (CFAR), A*STAR. The authors declare that they have no competing interests.

\bibliography{aaai23}

\renewcommand{\thesection}{S\arabic{section}}
\renewcommand{\thefigure}{S\arabic{figure}}
\renewcommand{\thetable}{S\arabic{table}}
\setcounter{figure}{0}
\setcounter{section}{0}
\setcounter{table}{0}

\section{Appendix}

\subsection{Experimental Details}

For experiments on both ciFAIR-10 and ciFAIR-100, PyTorch’s default implementation of cross entropy loss was used for object classification tasks. The SGD algorithm was used as the optimizer. The learning rate was set at a constant of 0.001. Momentum was fixed at 0.9. A batch size of 32 is used.

For ciFAIR-10, we employ a 2-layer 2D-convolutional network with 6 and 16 channels in the successive layers, followed by 3 fully connected layers with 400, 120 and 84 hidden units respectively. ReLU was used as the activation function. 

We follow the standard training and testing data splits from the original ciFAIR-10. 
In every task, the model is trained for 250 epochs. Each experiment is conducted with 20 runs initialized with 20 random seeds, where the seeds are used to vary the controlled variables. The average performance across all 20 runs is reported.

For ciFAIR-100, PyTorch's implementation of MobileNetV3 (small) was used, including the default layers and activation function. We used a custom training, validation, and test data splits with a ratio of 9:1:2, and a stopping criteria for training depending on the validation loss. The ciFAIR-100 images were upscaled to 72x72 using PyTorch's bicubic interpolation function before training.

\subsection{More Results and Analysis}
\begin{figure*}[h!]
\centering
\includegraphics[width=0.9\linewidth]{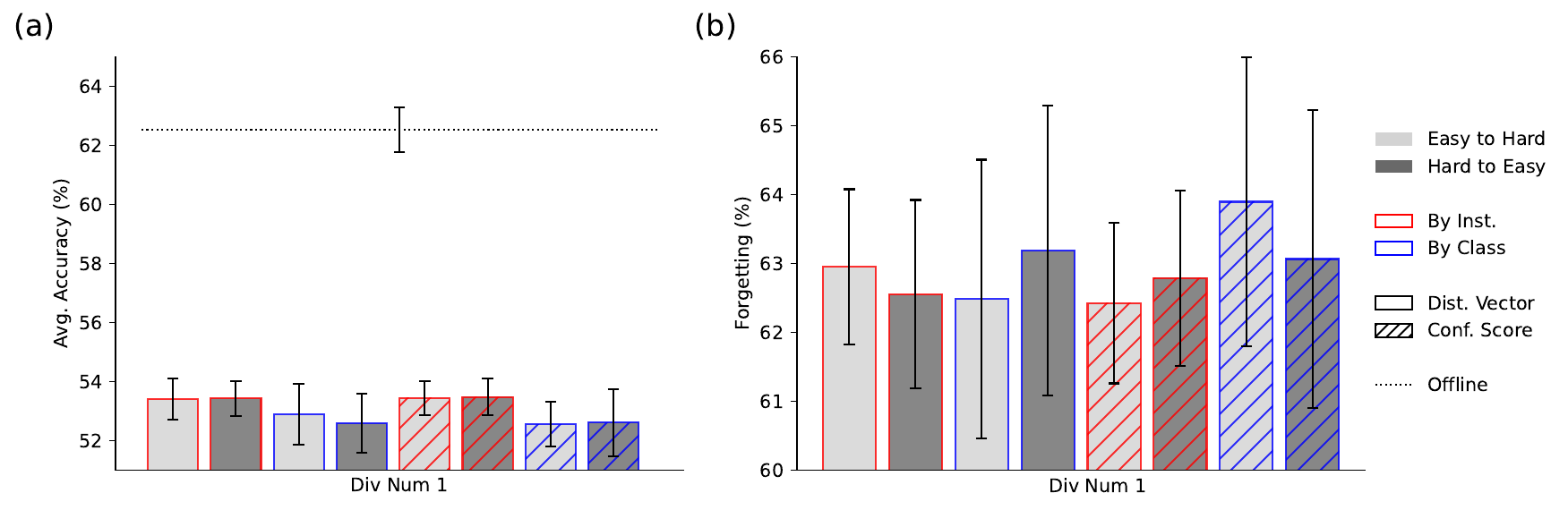}
\caption{\textbf{Continual learning performance on ciFAIR-10 dataset under different rehearsal sequences.} Average classification accuracy and Forgetfullness scores (F) over all tasks for models trained with different rehearsal sequences are reported. The higher the average accuracy, the better. The lower the F, the better. The design conventions follow the ones in Figure \ref{fig:exp2}.
}
\label{fig:exp2-cifar10}    
\end{figure*}

\begin{figure*}[h!]
\centering
\includegraphics[width=0.9\linewidth]{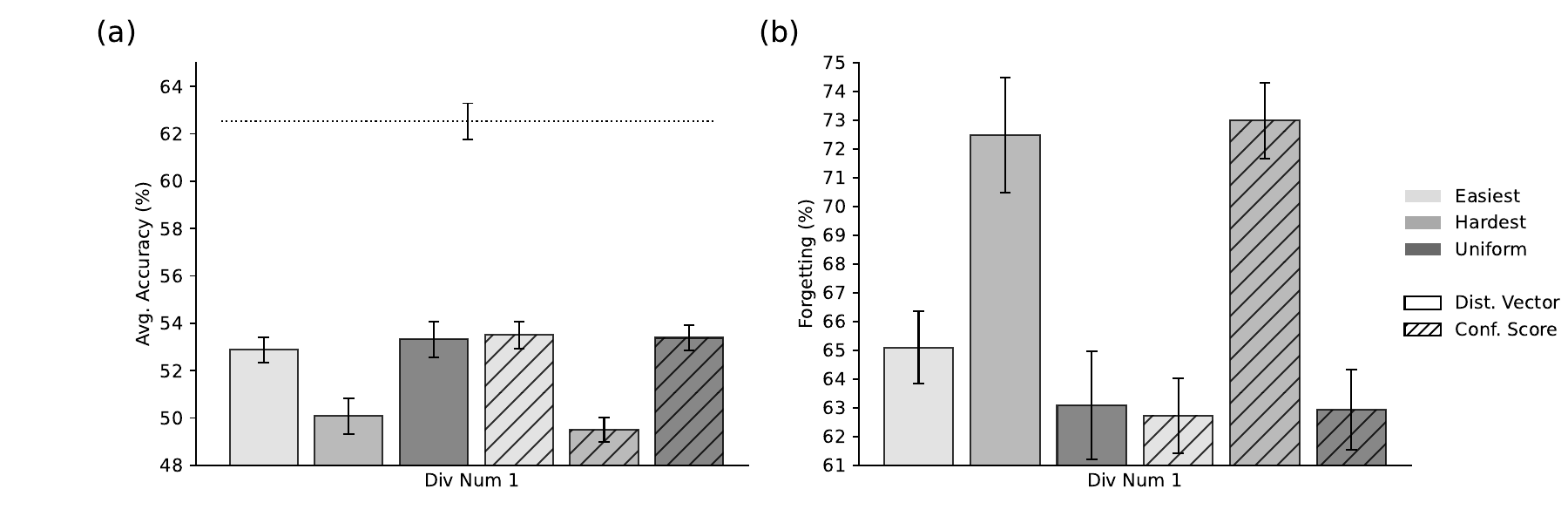}
\caption{\textbf{Continual learning performance on ciFAIR-10 dataset under different replay sample selection strategies.} Average classification accuracy and Forgetfullness scores (F) over all the tasks for models trained with different rehearsal samples based on different selection strategies are reported. The higher the average accuracy, the better. The lower the F, the better. The design conventions follow the ones in Figure \ref{fig:exp3}.
}
\label{fig:exp3-cifar10}    
\end{figure*}

We reported the continual learning performance on ciFAIR-10 dataset of the models trained with the three types of curricula as elaborated in \textbf{Experiments Section}.
See Table \ref{tab1:exp1-cifar10} for interleave divisions, Figure \ref{fig:exp2-cifar10}  for rehearsal sequences, and Figure \ref{fig:exp3-cifar10} for sample selections.
All the tables and figures on ciFAIR-10 dataset follow the same design conventions as the corresponding tables and figures on ciFAIR-100 dataset in the main text. The conclusions from the results of ciFAIR-10 dataset are consistent with the ones on the ciFAIR-100 dataset.

\begin{table}[h!]
\centering
\begin{tabular}{|c|c|c|c|c|c|}
\hline
Inter. Div. & 1    & 8    & 50   & 200  & 400  \\ \hline
F           & 62.5 & 59.3 & \textbf{57.1} & 57.4 & 57.2 \\ \hline
Avg.Accu    & 53.5 & 54.9 & \textbf{55.8} & \textbf{55.8} & \textbf{55.8} \\ \hline
\end{tabular}
\caption{\textbf{Continual learning performance on ciFAIR-10 as a function of interleave divisions (Inter. Div.)}. The higher Avg. Accu., the better; the lower F, the better. The best are in bold. See \textbf{Experiments Section} for evaluation metrics and experiments in interleave divisions.}
\label{tab1:exp1-cifar10}\vspace{-4mm}
\end{table}

\begin{figure*}[h!]
\centering
\includegraphics[width=0.9\linewidth]{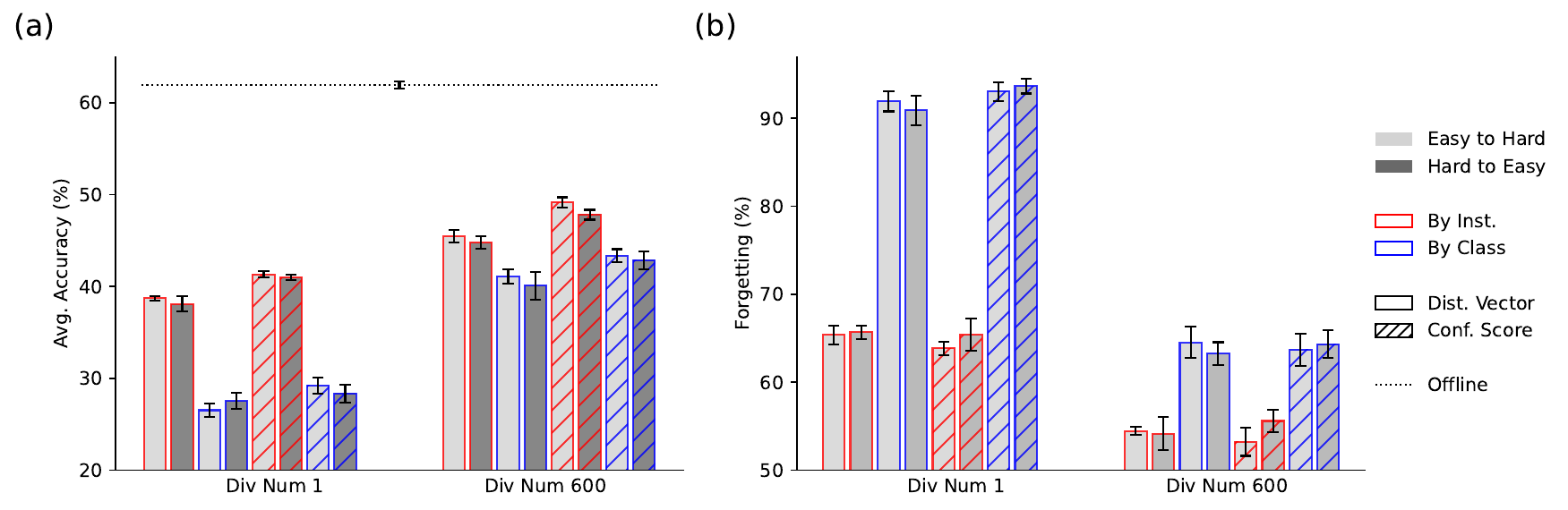}
\caption{\textbf{Continual learning performance on ciFAIR-100 dataset under different rehearsal sequences.} Complementary to the average accuracy results in Figure \ref{fig:exp2}, we also report the results of F scores on the ciFAIR-100 dataset. The results of Avg. Accu. are duplicated from Figure \ref{fig:exp2} and are also shown here again for completeness.
}
\label{fig:exp2-cifar100}    
\end{figure*}

\begin{figure*}[h!]
\centering
\includegraphics[width=0.9\linewidth]{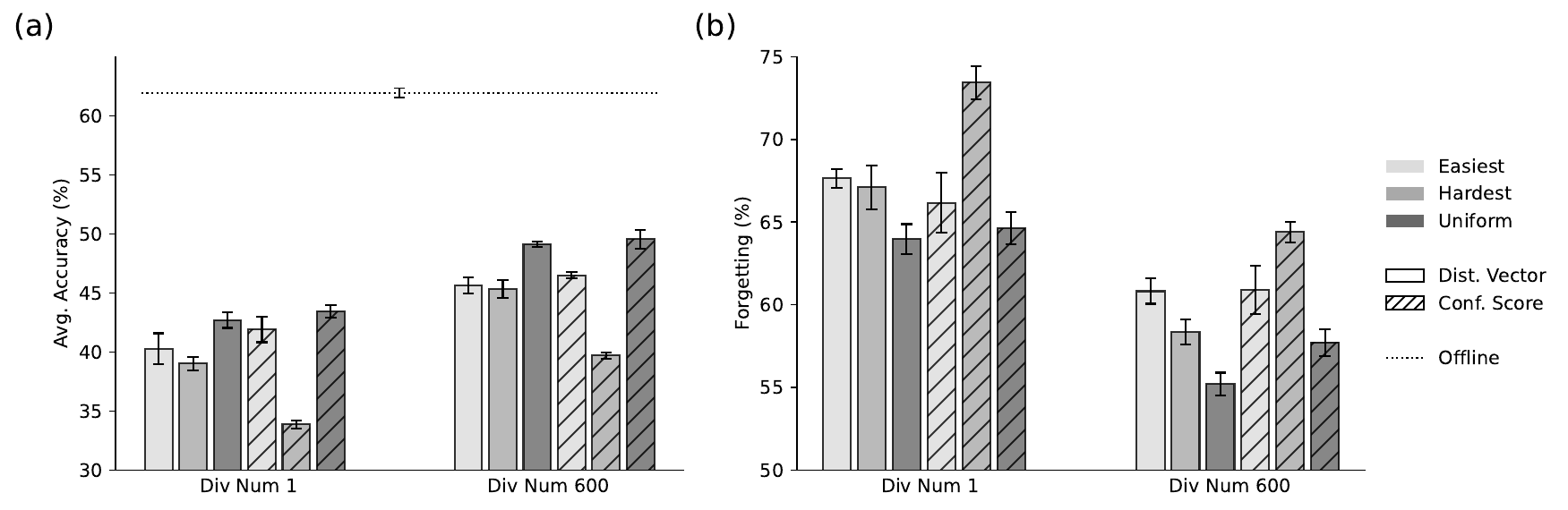}
\caption{\textbf{Continual learning performance on ciFAIR-100 dataset under different replay sample selection strategies.} Complementary to the average accuracy results in Figure \ref{fig:exp3}, we also report the results of F scores on the ciFAIR-100 dataset. The results of Avg. Accu. are duplicated from Figure \ref{fig:exp3} and are also shown here again for completeness. 
}
\label{fig:exp3-cifar100}    
\end{figure*}

\end{document}